\documentclass[sigconf]{acmart}

\usepackage{booktabs} % For formal tables
\usepackage{bm}
\usepackage{comment}
\usepackage{soul}
\usepackage{caption}
\usepackage{float}
\usepackage{amsfonts}
\usepackage[normalem]{ulem}
\usepackage[]{algorithm2e}
\usepackage{todonotes}
\usepackage{multirow}
\usepackage{subcaption,siunitx,booktabs}
\usepackage[]{fancyhdr}
\usepackage{diagbox}
\usepackage{makecell}

\usepackage{blindtext}

\AtBeginDocument{%
  \providecommand\BibTeX{{%
    \normalfont B\kern-0.5em{\scshape i\kern-0.25em b}\kern-0.8em\TeX}}}

\copyrightyear{2019} 
\acmYear{2019} 
\setcopyright{acmcopyright}
\acmConference[KDD '19]{The 25th ACM SIGKDD Conference on Knowledge Discovery and Data Mining}{August 4--8, 2019}{Anchorage, AK, USA}
\acmBooktitle{The 25th ACM SIGKDD Conference on Knowledge Discovery and Data Mining (KDD '19), August 4--8, 2019, Anchorage, AK, USA}
\acmPrice{15.00}
\acmDOI{10.1145/3292500.3330783}
\acmISBN{978-1-4503-6201-6/19/08}

\begin{document}

\title{Predicting Different Types of Conversions \protect\\ with Multi-Task Learning in Online Advertising}


\author{Junwei~Pan}
\affiliation{Yahoo Research}
\email{jwpan@verizonmedia.com}
\author{Yizhi~Mao}
\affiliation{Yahoo Inc.}
\email{lolam@verizonmedia.com}
\author{Alfonso~Lobos~Ruiz}
\affiliation{UC Berkeley}
\email{alobos@berkeley.edu}
\author{Yu~Sun}
\affiliation{Indeed}
\email{sunyu@indeed.com}
\author{Aaron~Flores}
\affiliation{Yahoo Research}
\email{aaron.flores@verizonmedia.com}

\renewcommand{\shortauthors}{J. Pan et al.}

\begin{abstract}
Conversion prediction plays an important role in online advertising since Cost-Per-Action (CPA) has become one of the primary campaign performance objectives in the industry. Unlike click prediction, conversions have different types in nature, and each type may be associated with different decisive factors.
In this paper, we formulate conversion prediction as a multi-task learning problem, so that the prediction models for different types of conversions can be learned together. These models share feature representations, but have their specific parameters, providing the benefit of information-sharing across all tasks. We then propose Multi-Task Field-weighted Factorization Machine (MT-FwFM) to solve these tasks jointly. Our experiment results show that, compared with two state-of-the-art models, MT-FwFM improve the AUC by 0.74\% and 0.84\% on two conversion types, and the weighted AUC across all conversion types is also improved by 0.50\%.

\end{abstract}

\begin{CCSXML}
<ccs2012>
<concept>
<concept_id>10010147.10010257.10010293.10010309</concept_id>
<concept_desc>Computing methodologies~Factorization methods</concept_desc>
<concept_significance>500</concept_significance>
</concept>
<concept>
<concept_id>10002951.10003227.10003447</concept_id>
<concept_desc>Information systems~Computational advertising</concept_desc>
<concept_significance>500</concept_significance>
</concept>
<concept>
<concept_id>10003752.10010070.10010099.10011253</concept_id>
<concept_desc>Theory of computation~Computational advertising theory</concept_desc>
<concept_significance>500</concept_significance>
</concept>
</ccs2012>
\end{CCSXML}

\ccsdesc[500]{Computing methodologies~Factorization methods}
\ccsdesc[500]{Information systems~Computational advertising}
\ccsdesc[300]{Theory of computation~Computational advertising theory}

\keywords{Online Advertising, Conversion Prediction, Factorization Machines, Multi-Task Learning}

\maketitle

\section{Introduction}
Online advertising is a 27.5-billion dollar business in fiscal year 2017~\cite{iab-report}, and advertisers have been shifting their budgets to programmatic ad buying platforms. Recently, more and more advertisers are running campaigns with Cost-Per-Action (CPA) goals, seeking to maximize conversions for a given budget. To achieve such objectives, accurate prediction of conversion probability is fundamental and has attracted lots of research attention in the past few years~\cite{lee2012estimating, agarwal2010estimating, chapelle2015simple,lu2017practical,ahmed2014scalable}. 

Advertising platforms insert \textit{pixels} (\textit{i.e.} Javascript codes) into advertisers' websites to track users' conversions, and there are several types of conversions that the advertisers want to track. Some pixels track whether a user fills out an online form, while other pixels track whether a user buys a product. The existence of different types of conversions makes conversion prediction challenging because the decisive factors that drive users to convert may vary from one conversion type to another. For example, whether to fill out a form online is a personal decision, so field \textit{User\_ID} and its interaction effects with other fields, should be the decisive factor. While for online purchase, the product itself or its corresponding brand play more important roles.

To address this problem, one approach is to build a separate model for each conversion type. However, this is memory intensive and it fails to leverage information from other conversion types. Another approach is to build a unified model which captures the 2-way or 3-way interactions between fields, with conversion type included as one of the fields. However, the 2-way model fails to capture the different field interaction effects for different conversion types, while the 3-way model is computationally expensive.

In this paper we study an alternative approach, \textit{i.e.}, formulating conversion prediction as a multi-task learning problem, so that we can jointly learn prediction models for multiple conversion types. Besides task-specific parameters, these models share low level feature representations, providing the benefit of information sharing among different conversion types. We propose Multi-Task Field-weighted Factorization Machine (MT-FwFM), based on one of the best-performing models for click prediction, \textit{i.e.}, Field-weighted Factorization Machine (FwFM)~\cite{pan2018field}, to solve these tasks together.

Our main contribution is two-fold: First, we formulate conversion prediction as a multi-task learning problem and propose MT-FwFM to solve all tasks jointly. Second, we have carried out extensive experiments on real-world conversion prediction data set to evaluate the performance of MT-FwFM against existing models. The results show that MT-FwFM increases the AUC of ROC on two conversion types by 0.74\% and 0.84\%, respectively. The weighted AUC of ROC across all tasks is also increased by 0.50\%. We have also conducted comprehensive analysis, which shows that MT-FwFM indeed captures different decisive factors for different conversion types.

The rest of the paper is organized as follows. We investigate  the field interaction effects for different conversion types in Section~\ref{sec:main_interaction_effect}. Section~\ref{sec:mtfwfm} describes MT-FwFM in detail. Our experiment results are presented in Section~\ref{sec:experiment}. In Section~\ref{sec:analysis}, we conduct analysis to show that MT-FwFM learns different field interaction effects for different conversion types. Section~\ref{sec:related_work} and Section~\ref{sec:conclusion} discuss the related work and conclude the paper.

\label{sec:introduction}

\section{Field Interaction Effects for Different Conversion Types}
The data used for conversion prediction are typically \emph{multi-field categorical data}~\cite{zhang2016deep}, where features are very sparse and each feature belongs to only one \emph{field}. For example, feature \textit{yahoo.com} and \textit{Nike} belong to field \textit{Page\_TLD} (Top-level domain) and \textit{Advertiser}, respectively. In click prediction, it has been verified that different field pairs have different interaction effects on multi-field categorical data~\cite{juan2016field, pan2018field}.

%\emph{field interaction effects}

In conversion prediction, advertisers would like to track different types of conversions, and they spend most of their budget on the following four types:
\begin{itemize}
    \item \textit{Lead}: the user fills out an online form
    \item \textit{View Content}: the user views a web page such as the landing page or a product page
    \item \textit{Purchase}: the user purchases a product
    \item \textit{Sign Up}: the user signs up an account
\end{itemize}
%\textit{Lead}, \textit{View Content}, \textit{Purchase}, and \textit{Sign Up}. \textit{Lead} refers to filling out an online form; \textit{View Content} denotes whether a key web page such as the landing page or a product page is viewed by a user; \textit{Purchase} refers to purchasing a product; \textit{Sign Up} refers to signing up an account. 

%Thus, we conduct a comprehensive analysis in this section to verify that the main and interaction effects indeed differ among different conversion types.

\begin{table*}
\centering
	\begin{tabular}{| c | c |}
    \hline
    Conversion Type & Top 5 Field Pairs \\
    \hline
	Lead  & (Ad, User), (Creative, User), (Line, User), (Subdomain, User), (Advertiser, User) \\
    \hline
    View Content  & (Subdomain, Hour), (Ad, Subdomain), (Creative, Subdomain), (Subdomain, Age\_Bucket), (Page\_TLD, Hour) \\
    \hline
    Purchase  &  (Ad, Subdomain), (Creative, Subdomain), (Ad, Page\_TLD), (Creative, Page\_TLD), (Line, Subdomain)\\
    \hline
	Sign Up & (Ad, Subdomain), (Creative, Subdomain), (Ad, Age\_Bucket), (AD, Page\_TLD), (Creative, Page\_TLD)\\
    \hline
	\end{tabular}
    \caption{Top 5 field pairs in terms of mutual information for each conversion type. Please refer to Section~\ref{subsec:data set} for the description of these fields in detail.}
    \label{table:field_interaction}
\end{table*}

%The data set can be denoted as $\mathcal{S} = \{(\bm{x},y,t)\}$, where for each sample, $\bm{x} \in \{0,1\}^m$ is the feature vector with $x_i = 1$ for active features, 

The decisive factors, \textit{i.e.}, the main effect terms (fields) and/or the interaction terms (field pairs) that drive a user to convert, may vary a lot among these types. Following the analysis in~\cite{pan2018field}, we verify this by computing mutual information (MI) between each field pair and each type of conversion on our real-world data set described later in section~\ref{subsec:data set}. Suppose there are $M$ unique features $\{x_1,\dots,x_M\}$, $N$ different fields $\{F_1,\dots,F_N\}$ and $T$ conversion types. We denote $F(i)$ as the field that feature $i$ belongs to, and  $t \in \mathcal{T}$ as the conversion type. The interaction effect of a field pair $(F_p, F_q)$ with respect to conversions of type $t$ is measured by:

\begin{equation}\label{eq:mi}
MI^t((F_p,F_q), Y) = \sum_{(i,j) \in (F_p, F_q)} \sum_{l \in \{0,1\}} p^t((i,j), l) \log \frac{p^t((i,j), l)}{p^t(i,j)p^t(l)}
\end{equation}

\noindent where $p^t((i,j),l)$ is the marginal probability of $p^t(x_i=1,x_j=1,y=l)$, $p^t(i,j)$ denotes $p^t(x_i=1,x_j=1)$, and $p^t(l)$ is the marginal probability of $p^t(y=l)$. All marginal probabilities are computed based on the samples from each conversion type $t$. 

The top 5 field pairs that have the highest mutual information \textit{w.r.t.} each conversion type are shown in Table~\ref{table:field_interaction}. It shows that these field pairs vary among types: all 5 field pairs of \textit{Lead} contain field \textit{User\_ID} and all 5 field pars of \textit{View Content} contain publisher fields (\textit{Page\_TLD} and \textit{Subdomain})\footnote{\textit{Page\_TLD} denotes a top-level domain of a web page, while \textit{Subdomain} denotes the subdomain. For example, given a web page with URL \textit{https://sports.yahoo.com/warriors-loss-76ers-vivid-illustration-075301147.html}, the \textit{Page\_TLD} is \textit{yahoo.com} and the \textit{Subdomain} is \textit{sports.yahoo.com}}. For \textit{Purchase} and \textit{Sign Up}, most field pairs contain one publisher field and one advertiser field (\textit{Ad}, \textit{Creative}, \textit{Line}). The heat maps of the mutual information for all field pairs with respect to each conversion type are shown in Figure~\ref{fig:heat-map-mi-different-conv-type} and please refer to section~\ref{subsec:data set} for the explanation of each field.

There are several approaches to capture different field interaction effects for different conversion types. The first one is to build one model for each conversion type, and train each model separately. However, this is not preferred in the real-world advertising platform because lots of memories are required to store the parameters of all models. In addition, extreme low conversion rate for some conversion types may render the lack of sufficient positive samples to train the corresponding models.

The second approach is to build a unified model, with conversion type as one of the fields. However, all 2-way state-of-the-art models, such as 2-way Factorization Machines (FM) and Field-weighted Factorization Machines (FwFM), are not able to fully capture the differences in field interaction effects among different conversion types. 3-way FM and FwFM may resolve this issue, but the online computing latency is much higher. Please refer to Section~\ref{subsubsec:computing_time} for the details.
\label{sec:main_interaction_effect}

\section{Multi-task Field-weighted Factorization Machine}
\label{sec:mtfwfm}
We formulate the prediction of different types of conversions as a multi-task learning problem, and propose Multi-Task Field-weighted Factorization Machine (MT-FwFM) to train these models jointly. This section is organized as follows: Section~\ref{subsec:mt-fwfm} introduces FwFM and MT-FwFM in detail; the training procedure of MT-FwFM is described in section~\ref{subsec:jointtrain}. In Section~\ref{subsec:modelcomplex}, we analyze the number of parameters as well as computing latency for MT-FwFM. 

\subsection{Multi-Task Field-weighted Factorization Machine (MT-FwFM)}\label{subsec:mt-fwfm}

MT-FwFM is a variant of Field-weighted Factorization Machine (FwFM), which is introduced in~\cite{pan2018field} for click prediction. FwFM is formulated as

\begin{equation}\label{eq:fwfm1}
p(y|\Theta, \bm{x}) = \sigma(\Phi_{FwFM}(\Theta, \bm{x}))
\end{equation}

\noindent where $\sigma(x)$ is the sigmoid function, and $\Phi_{FwFM}(\Theta, \bm{x})$ is the sum of the main and interaction effects across all features:

\begin{equation}\label{eq:fwfm2}
\begin{split}
	\Phi_{FwFM}(\Theta, \bm{x}) &= w_0 + \sum_{i=1}^M x_i \langle \bm{v}_i, \bm{w}_{F(i)}\rangle \\
    &+ \sum_{i=1}^M\sum_{j=i+1}^M x_{i} x_{j} \langle \bm{v}_i, \bm{v}_j \rangle r_{F(i), F(j)}
\end{split}
\end{equation}

\noindent Here $\Theta$ is a set of parameters $\{w_0, \bm{w}, \bm{v}, r\}$: $w_0$ denotes the bias term; $\bm{v}_i$ refers to the embedding vector for feature $i$;  $\bm{w}_{F(i)}$ denotes \emph{main term weight vector} for field $F(i)$, which is used to model the main effect of feature $i$; $r_{F(i), F(j)}$ denotes the \emph{field interaction weight} between field $F(i)$ and $F(j)$.  

We modify FwFM in the following ways to get MT-FwFM: First, instead of using one bias term $w_0$, MT-FwFM has one bias term $w_0^t$ for each conversion type $t$. Second, each conversion type has its own $\bm{w}_{F(i)}^t$ to model the main effect of feature $i$. Last, each conversion type also has its own field interaction weights $r_{F(i), F(j)}^t$. The feature embeddings $\bm{v}_i$ are kept the same as FwFM and are shared by all conversion types. Mathematically,

\begin{equation}
\begin{split}
\Phi_{MT-FwFM}(\Theta, \bm{x}) &= w_0^{t} + \sum_{i=1}^M x_i \langle \bm{v}_i, \bm{w}_{F(i)}^{t} \rangle \\
&+ \sum_{i=1}^M\sum_{j=i+1}^M x_i x_j \langle\bm{v}_i, \bm{v}_j \rangle r_{F(i), F(j)}^{t}
\end{split}
\end{equation}

MT-FwFM can be regarded as a 3-layer neural network: each sample is first processed by an \textit{embedding layer} that maps each binary feature $x_i$ to an embedding vector $\bm{v}_i$, then by a \textit{main \& interaction layer} which consists of $\bm{v}_i$ and $\langle \bm{v}_i, \bm{v}_j \rangle$. 

\begin{figure}[ht]
\centering
    \includegraphics[scale=0.34]{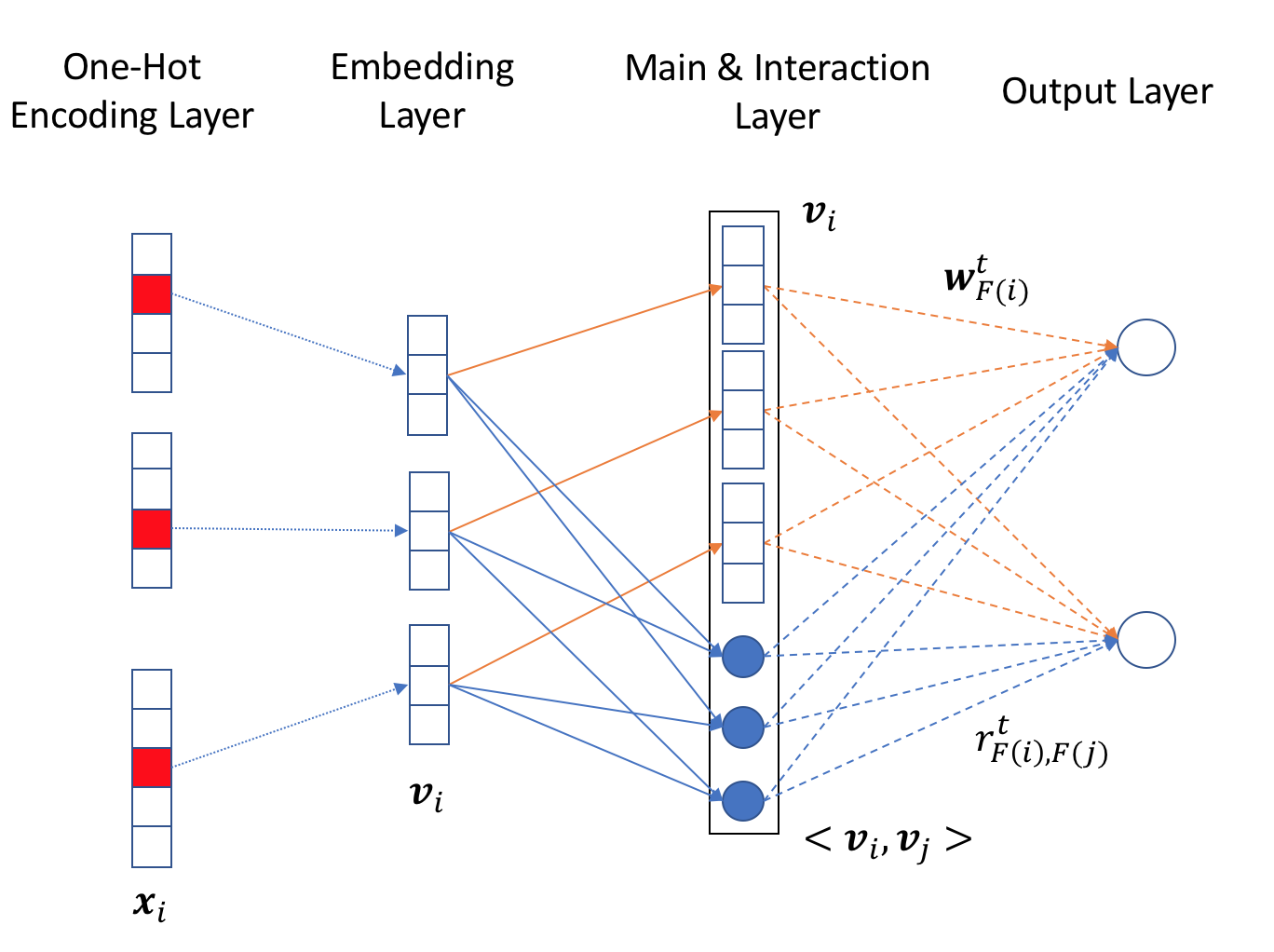}
    \caption{Architecture of MT-FwFM. In the one-hot encoding layer, each field has only one active feature, represented as the red squares. Each active feature is then mapped to an vector in the embedding layer. The blue rectangles in main \& interaction layer are copies of the vectors from the embedding layer, while the blue circles represent the dot products between embedding vectors, \textit{i.e.}, $\langle \bm{v}_i, \bm{v}_j\rangle$. There are $T$ nodes in the output layer, one for each conversion type. The orange dash lines that connect $\bm{v}_i$ with each output node are weighted by the main term weight vectors $\bm{w}^t_{F(i)}$, while the connections between blue circles and each output node are weighted by the field interaction weights $r_{F(i), F(j)}^t$. We omitted $w_0^t$ in this figure for the sake of simplicity.}
    \label{fig:tensorflow}
\end{figure}

Each node in the main and interaction layer is connected to a \textit{output layer} which consists of $T$ nodes, one for each conversion type. The connections between $\bm{v}_i$ and each output node are weighted by $\bm{w}_{F(i)}^t$, while connections between $\langle \bm{v}_i, \bm{v}_j \rangle$ and each output node are weighted by field interaction weights $r_{F(i),F(j)}^t$. The architecture of MT-FwFM is shown in Figure~\ref{fig:tensorflow}.

\subsection{Joint Training} \label{subsec:jointtrain}

The feature embedding vectors $\bm{v}_i$ are shared during the model training by all conversion types and are optimized for every sample. However, for the conversion type specific parameters such as $w_0^t, \bm{w}_{F(i)}^t$ and $r_{F(i), F(j)}^t$, they are only optimized for samples of corresponding type. We minimize the following loss function for MT-FwFM:

%Even though there are \textcolor{red}{$T$} nodes, one for each conversion type, in the output layer of MT-FwFM. For a given training sample $\bm{x}$, only the output node that corresponds to its conversion type $t$ is active. Therefore we optimize the loss function w

%and we optimize the logistic loss for that output node only.

%In MT-FwFM, even though there are \textcolor{red}{$T$} nodes for each conversion type in the output layer, for a given training instance $\bm{x}$, only the one corresponding to its conversion type $t$ will be active, and we optimize the logistic loss for that output node only. The other output nodes cause no loss since they are not active for this instance. Mathematically, the loss function is 

\begin{equation}
    \sum_i -y_i \log \hat{y_i} - (1-y_i) \log (1-\hat{y_i}) + \lambda \Omega(\Theta)
\end{equation}

\noindent where $\hat{y}_i=p(y|\Theta, \bm{x}^{(i)})$, $y_i$ denotes the label, and $\Omega(\Theta)$ denotes the regularization terms \textit{w.r.t.} the parameters. 

We use mini-batch stochastic gradient descent to optimize the loss function. In each iteration, we select a batch of samples($\mathcal{B}$) randomly, where each sample belongs to a specific task, \textit{i.e.}, conversion type in our case. Within each batch, the model is updated according the conversion type of each sample. More specifically, $\bm{v}_i$ is updated for all samples, while $w_0^t$, $\bm{w}^t_{F(i)}$ and $r^t_{F(i),F(j)}$ are updated only for samples with conversion type $t$. The training procedure is summarized in Algorithm~\ref{alg:mt_fwfm}. 

\begin{algorithm}
\SetAlgoLined
\KwData{$\mathcal{S}=\{(\bm{x},y,t)\}$}
\ Initialize parameter $\Theta:\{w,\bm{v},\bm{w},r\}$ randomly \;
\For{$epoch=1$ \KwTo $\infty$}{
    	Sample a set of training samples $\mathcal{B}$ from $\mathcal{S}$ randomly
        
        Compute log loss: $L(\Theta)= \sum_{(\bm{x},y,t) \in \mathcal{B}}-y \log \hat{y} - (1-y) \log (1-\hat{y}) + \lambda \Omega(\Theta)$
        
        Compute gradient: $\nabla(\Theta)$
        
        Update model: $\Theta = \Theta - \eta \nabla(\Theta)$
    
}
\caption{Training procedure of MT-FwFM.}
\label{alg:mt_fwfm}
\end{algorithm}

\subsection{Model Complexity} \label{subsec:modelcomplex}

There are two key constraints when we build a conversion prediction model in the real-time serving system: the memory needed to store all parameters, and the computing latency for each sample. We'll analyze these two constraints in this section.

%There are two key constraints when we build a conversion prediction model in the production environment of our advertising platform: the memory amount to store all parameters, and the online computing latency for each incoming prediction request.

\subsubsection{Number of Parameters}

The number of parameters in MT-FwFM is 
\begin{equation}\label{eq:num_parafwfm1}
T + MK + NTK + \frac{N(N-1)}{2}T \approx MK
\end{equation}
\noindent where $T$, $M$, $N$, $K$ refer to the number of conversion types, features, fields, as well as the dimension of the feature embedding vectors and main term weight vector, respectively. 

Thus in~(\ref{eq:num_parafwfm1}), $T$ represents the number of bias terms $w_0^t$; $MK$ calculates the number of parameters for the embedding vectors $\bm{v}_i$; $NTK$ corresponds to $\bm{w}_{F(i)}^t$, \textit{i.e.}, the main term weight vectors for all conversion types; $\frac{N(N-1)}{2}T$ denotes the number of field interaction weights $r_{F(i),F(j)}^t$. The number of parameters approximately equals to $MK$, given that $T \ll M$ and $N \ll M$. 

\subsubsection{Online Computing Latency}

The online computing latency for each prediction request grows linearly with the number of operations, such as float additions and multiplications. During the inference of MT-FwFM, for each sample, the number of operations in the main effect terms is $$N \cdot (2K-1) + (N-1)$$ and the number of operations in the interaction terms is $$\binom{N}{2} \cdot 2K + \binom{N}{2}-1$$

Thus the total number of operations  of MT-FwFM is 
$$N^2K + NK + \binom{N}{2} \approx N^2K$$

%\footnote{Total number of operations = $N \cdot (2K-1) + (N-1) + \binom{N}{2} \cdot 2K + \binom{N}{2} -1 + 2 = 2NK + \binom{N}{2}(2K+1) = N^2K + NK + \binom{N}{2}$}

\subsubsection{MT-FwFM \textit{v.s.} Using Conversion Type as a Field} \label{subsubsec:computing_time}

Besides formulating conversion prediction as a multi-task learning problem, an alternative approach is to incorporate conversion type as one of the fields in the existing models, such as FM and FwFM. We can either consider the 2-way interactions between fields, referred as \emph{2-way Conversion Type as a Field} (2-way CTF), or the 3-way interactions, referred as \emph{3-way Conversion Type as a Field} (3-way CTF). 2-way CTF with FM and FwFM are used as baseline models in Section~\ref{sec:experiment}.

For 3-way CTF with FM or FwFM, the number of operations is much more than that of MT-FwFM, which makes them less preferred in the production environment. We discuss the number of operations of 3-way CTF with FwFM as an example here and omit that for FM since they are very similar. The formula of 3-way CTF with FwFM are:

\begin{equation}
\begin{split}
\Phi(\Theta, \bm{x}) &= w_0 + \sum_{i=1}^{M+T}x_i \langle \bm{v}_i, \bm{w}_{F(i)} \rangle \\
&+ \sum_{i=1}^{M+T} \sum_{j=i+1}^{M+T} \langle \bm{v}_i, \bm{v}_j \rangle \\
&+ \sum_{i=1}^M \sum_{j=i+1}^M \sum_{t=1}^T x_i x_j x_t \langle \bm{v}_i, \bm{v}_j, \bm{v}_t \rangle r_{F(i), F(j)}
\end{split}
\end{equation}

\noindent where $\langle \bm{v}_i, \bm{v}_j, \bm{v}_t \rangle = \sum_{k=1}^K v_i^{(k)} \cdot v_j^{(k)} \cdot v_t^{(k)} $ is a 3-way dot product.

The number of operations of 3-way CTF with FwFM is 

\begin{equation}
    \Big(\frac{5}{2}N^2 + \frac{3}{2}N+2\Big)K + \binom{N}{2}
\end{equation}

% \begin{equation}
%     \binom{N}{2}(3K+1) + \binom{N+1}{2}2K + (N+1)(2K-1) + N + 1
% \end{equation}

% \begin{equation}
%     (N+1)(2K-1) + N + \binom{N+1}{2}2K + \binom{N}{2}(3K+1) + 1
% \end{equation}

It is approximately $\frac{5}{2}N^2K$, which is 150\% more than that of MT-FwFM. Thus, compared with MT-FwFM, 3-way CTF with FwFM is less preferred due to its much more number of operations.

\section{Experiments}
\label{sec:experiment}
This section presents our experimental evaluation results. We introduce the data set in Section~\ref{subsec:data set}, and describe the implementation details in Section~\ref{subsec:implementation}. Section~\ref{subsec:performance} compares the performance of MT-FwFM with that of 2-way CTF with FM and FwFM. We denote 2-way CTF with FM or FwFM as FM or FwFM in this section for the sake of simplicity.

\subsection{Data Set}
\label{subsec:data set}

The data set is collected from the impression and conversion logs of the Verizon Media DSP advertising platform. We treat each impression as a sample, and use the conversions to label them. The labeling is done by \emph{last-touch attribution}, \text{i.e.}, for each conversion, only the last impression(from the same user and \emph{line} \footnote{Line is the smallest unit for advertisers to set up budget, goal type, targeting criteria of a group of ads}) before this conversion is labeled as a positive sample. All the remaining impressions are labeled as negative samples. The type of each sample is the type of the corresponding line. A line may be associated with multiple conversions that belong to several different types. However, in this paper we focus on those lines that have only one type of conversions since they contribute to most of the traffic as well as spend in our platform.

%All impressions with conversions attributed to them by last-touch attribution are labeled as positive samples. The conversion types of these impressions are the types of the attributed conversions. All remaining impressions without any conversions attributed to are labeled as negative samples. The type of these impressions are the type of conversions of the corresponding campaigns. A campaign may have multiple conversions that belong to several different types. However, in this section we focus on those campaigns that have only one type of conversions since they contribute to most of the traffic as well as revenue in our platform.

We use 7 days of impression logs, denoted as $T_1$ to $T_7$, as the training data set. Then conversions from $T_1$ to $T_{13}$ are used to label those impressions. A 6-days longer conversion time window is used because there are usually delays between impressions and conversions, and most conversions happens within 6 days after impressions. We then downsample the negative samples to solve the data imbalance issue since the ratio of positive samples is in the order of $10^{-4}$ in the data set. We get approximately equal number of positive and negative samples in the training set after downsampling.

The validation data set is collected from the impression logs on $T_{8}$, and the test data set is collected on $T_{9}$. Conversions from $T_{8}$ to $T_{14}$ and $T_{9}$ to $T_{15}$ are used to label the validation and test set, respectively. We do not downsample on validation and test data sets, since the evaluation should be applied to data sets that reflect the real class distribution. Table~\ref{table:datasets} summarizes the statistics of the training, validation and test data set.

\begin{table}
\centering
    \begin{tabular}{ | c | c | r | c | c |} 
    \hline
    \multicolumn{2}{|c|}{Data set} & Samples & CVR &  Features \\
    \hline
    \multirow{4}{*}{Train} & Purchase & 4,552,380 & 0.1858 & 11,852\\
    \cline{2-5}
    & Lead & 6,566,688 & 0.3402 & 15,728 \\
    \cline{2-5}
    & Sign Up & 3,332,250 & 0.8797 & 13,227 \\
    \cline{2-5}
    & View Content & 170,694 & 0.3690 & 1,171 \\
    \hline
    \multirow{4}{*}{Validation} & Purchase & 12,800,160 &  4.63E-04 & 11,153 \\
    \cline{2-5}
    & Lead & 17,036,604 & 5.59E-04 & 9,474 \\
    \cline{2-5}
    & Sign Up & 2,222,334 & 3.30E-03 & 5,591 \\
    \cline{2-5}
    & View Content & 441,252 & 4.90E-04 & 1,494 \\
    \hline
    \multirow{4}{*}{Test} & Purchase & 12,623,382 & 4.52E-04 & 11,007 \\
    \cline{2-5}
    & Lead & 18,738,990 & 5.37E-04 & 9,373 \\
    \cline{2-5}
    & Sign Up & 1,926,558 & 3.41E-03 & 5,553 \\
    \cline{2-5}
    & View Content & 383,940 & 4.69E-04 & 1,173 \\
    \hline
 	\end{tabular}
    \caption{Statistics of training, validation and test data sets.}
    \label{table:datasets}
\end{table}

\begin{table*}
\centering
\sisetup{table-format=4.0} % integer values only, up to 4 digits
\begin{tabular}{| c | c | c | c | c | c | c | } 
\hline
	\multirow{2}{*}{Model} & \multicolumn{3}{c|}{Overall AUC}  & \multicolumn{3}{c|}{Weighted AUC} \\ \cline{2-7}
    & Training & Validation & Test & Training & Validation & Test \\
	%\hline
	%LR & 0.9665 & \textbf{0.9040} & 0.9032 & 0.9489 & 0.8501 & 0.8361 \\ % 45
    \hline
	FM & 0.9706 & 0.9014 & 0.9012 & 0.9537 & 0.8500 & 0.8383 \\ % 22
	\hline
	FwFM & 0.9702 & \textbf{0.9023} & 0.9027 & 0.9530 & \textbf{0.8520} & 0.8400 \\ % 31
    \hline
    MT-FwFM & \textbf{0.9728} & 0.8999 & \textbf{0.9046} & \textbf{0.9574} & 0.8511 & \textbf{0.8450} \\ % 120
    \hline
\end{tabular}
\caption{Performance comparison on real-world conversion data set.}
\label{table:mt-fwfm-vs-others}
\end{table*}

\begin{table*}
\centering
\sisetup{table-format=4.0} % integer values only, up to 4 digits
\begin{tabular}{| c | c | c | c | c |} 
\hline
	Type & Model & \makecell{Training  AUC}  & \makecell{Validation  AUC} & \makecell{Test  AUC} \\ 
	\hline
    \multirow{3}{*}{Lead} & FM & 0.8393  & 0.8412  & 0.8116 \\ \cline{2-5}
    & FwFM & 0.8357  & \textbf{0.8536}  & 0.8109 \\ \cline{2-5}
    & MT-FwFM & \textbf{0.8502}  &  0.8258 & \textbf{0.8190} \\ 
    \hline
	\multirow{3}{*} \makecell{View Content} & FM & 0.9523  & 0.9577  & 0.9542  \\ \cline{2-5}
    & FwFM & 0.9511 & 0.9569 & 0.9537 \\ \cline{2-5}
    & MT-FwFM & \textbf{0.9563} & \textbf{0.9580} & \textbf{0.9545} \\ 
    \hline
    \multirow{3}{*}{Purchase} & FM & 0.9922  & 0.9758 & 0.9684  \\ \cline{2-5}
    & FwFM & 0.9924 & 0.9804 & \textbf{0.9761} \\ \cline{2-5}
    & MT-FwFM & \textbf{0.9930} & 0.9799 & 0.9737 \\ 
    \hline
    \multirow{3}{*}{Sign Up} & FM & 0.9381 & 0.7529 & 0.7475 \\ \cline{2-5}
    & FwFM & 0.9374 & \textbf{0.7564} & 0.7501 \\ \cline{2-5}
    & MT-FwFM & \textbf{0.9428} & 0.7545 & \textbf{0.7585} \\ 
    \hline
\end{tabular}
\caption{Performance comparison on data set of each conversion type.}
\label{table:mtl-fwfm-vs-others-per-conversion-type}
\end{table*}

There are 17 fields of features, which fall into 4 categories: 
\begin{enumerate}
  \item User-side fields: \textit{User\_ID}, \textit{Gender} and \textit{Age\_Bucket}
  
  \item Publisher-side fields: \textit{Page\_TLD}, \textit{Publisher\_ID}, and \textit{Subdomain}
  
  \item Advertiser-side fields: \textit{Advertiser\_ID}, \textit{Creative\_ID}, \textit{AD\_ID}, \textit{Creative\_Media\_ID}, \textit{Layout\_ID}, and \textit{Line\_ID}
  
  \item Context fields: \textit{Hour\_of\_Day}, \textit{Day\_of\_Week}, \textit{Device\_Type\_ID}, \textit{Ad\_Position\_ID}, and \textit{Ad\_Placement\_ID}
\end{enumerate}

%(1) user side fields such as \textit{Gender}, \textit{Age\_Bucket}, and \textit{User\_ID}; (2) publisher side fields such as \textit{Page\_TLD}, \textit{Publisher\_ID}, and \textit{Subdomain}; (3) advertiser side fields such as \textit{Advertiser\_ID}, \textit{Ad\_ID}, \textit{Creative\_ID}, \textit{Creative\_Media\_ID}, \textit{Layout\_ID} , and \textit{Line\_ID}; (4) context side fields such as \textit{Hour\_of\_Day}, \textit{Day\_of\_Week}, \textit{Ad\_Position\_ID}, \textit{Ad\_Placement\_ID} and \textit{Device\_Type\_ID}. Specially, w
\noindent We use \textit{Conversion\_Type\_ID} as an additional field for FM and FwFM. The meanings of most fields are quite straightforward so we only explain some of them:
\begin{itemize}
    \item \textit{Page\_TLD}: top-level domain of a web page.
    \item \textit{Subdomain}: subdomain of a web page.
    %\item \textit{Line}: the smallest unit for advertisers to set up budget, goal type and targeting criteria of a group of ads.
    \item \textit{Creative\_ID}: identifier of a creative, which is an image or a video.
    \item \textit{Ad\_ID}: identifier of a (\textit{Line\_ID}, \textit{Creative\_ID}) combination.
    \item \textit{Creative\_Media\_ID}: identifier of the media type of the creative, \textit{i.e.}, image, video or native.
    \item \textit{Layout\_ID}: the size of a creative, for example, $300 \times 200$.
    \item \textit{Device\_Type\_ID}: identifier of whether this event happens on desktop, mobile or tablet.
    \item \textit{AD\_Position\_ID} \& \textit{AD\_Placement\_ID}: identifiers of the position of an ad on the web page.
\end{itemize}

\subsection{Implementations}
\label{subsec:implementation}

All baseline models as well as the proposed MT-FwFM model are implemented in Tensorflow. The input is a sparse binary vector $\bm{x} \in \mathbb{R}^M$ with only $N$ non-zero entries. In the embedding layer, the input vector $\bm{x}$ is projected into $N$ embedding vectors $\bm{v}_i$, one for each field. The main and interaction effect terms in the next layer, \textit{i.e.}, main \& interaction layer, are computed based on these $N$ vectors. The main effect terms simply concatenate all $N$ vectors, while the interaction effect terms calculate the dot product $\langle \bm{v}_i, \bm{v}_j \rangle$ between each feature pair. Then, each node in the main \& interaction layer is connected to the output layer, which consists of $T$ nodes, each of them corresponds to one specific conversion type. 

%While the embedding layer is shared by all tasks, the weights that connect the main \& interaction terms to each output node are separate.

\subsection{Performance Comparisons}
\label{subsec:performance}

This section compares MT-FwFM with FM and FwFM on the data sets introduced above. For the hyper-parameters such as regularization coefficient $\lambda$ and learning rate $\eta$ in all models, we select the values that lead to the best performance on the validation set and then use them in the evaluation on the test set. We focus on the following performance metrics:

\paragraph{\textbf{Overall AUC}} AUC of ROC (AUC) specifies the probability that, given one positive and one negative sample, their pairwise rank is correct. Overall AUC calculates the AUC over samples from all conversion types.

\paragraph{\textbf{AUC for each conversion type}} The AUC on the samples from each conversion type, denoted as $AUC_t$.

\paragraph{\textbf{Weighted AUC}} 

The weighted average of the AUC on each conversion type: $$\frac{\sum_{t \in \mathcal{T}} AUC_t \cdot N_t}{\sum_{t \in \mathcal{T}}N_t}$$
\noindent where $N_t$ refers to the spend of conversion type $t$. The weights $N_t$ are the spend of each conversion type.

%Two metrics are used to evaluate the performances of all models. The first one is the AUC of ROC on the whole data set across different conversion types, which is referred to as \emph{overall AUC}. For the second metric, we first get the AUC of ROC on the validation and test set for each conversion type, and then calculate a weighted average over all types. \textcolor{orange}{Total cost} (the revenue from advertiser) of each conversion type is used as the weights, so as to skew the metrics towards those conversion types with less samples but huge revenue impact. This weighted AUC of ROC is referred to \emph{weighted AUC}\footnote{Stratified AUC?}:

 Table~\ref{table:mt-fwfm-vs-others} summarizes the experiment results. It shows that MT-FwFM gets the best performance \textit{w.r.t.} both overall and weighted AUC, with a lift of $0.19\%$ and $0.50\%$ over the best performing baseline, respectively. While the performance improvement on overall AUC is marginal, the lift on weighted AUC is significant. 
 
Table~\ref{table:mtl-fwfm-vs-others-per-conversion-type} compares the performance of all models on each conversion type. Among four conversion types, \emph{View Content} and \emph{Purchase} have high AUCs than the other two types using the baseline models(over 95\% \textit{v.s.} under 82\%). For these two conversion types that already get high AUC, the lifts of MT-FwFM are more or less neutral, namely $0.03\%$ and $-0.24\%$. On the other hand, for conversion type \emph{Lead} and \emph{Sign Up} that get low performance on baseline models, MT-FwFM improves the AUC by $0.74\%$ and $0.84\%$.

%We observe that for the conversion types that already have high AUCs, such as \emph{View Content} and \emph{Purchase}, the lifts of MT-FwFM are small or even negative, namely $0.08\%$ and $-0.24\%$. On the other hand, for the conversion types with relatively low AUCs, namely \emph{Lead} and \emph{Sign Up}, the lifts of MT-FwFM are as \textcolor{orange}{significant} as $0.81\%$ and $0.84\%$. 
 
Therefore, we conclude that MT-FwFM outperforms FM and FwFM significantly \textit{w.r.t.} the weighted AUC over all conversion types. And this improvement mainly come from the conversion types that get relatively low AUC using the baseline models.

%We observe that, first, MT-FwFM get the best performance on the training data set w.r.t. both overall AUC of ROC and weighted AUC of ROC, which means that MT-FwFM can fit the training data best. Second, MT-FwFM get the best performance on the test data set w.r.t. both overall and weighted AUC of ROC, which proves that our proposed model generalizes the best to test data set. FwFM gets the second best weighted AUC on the test data set, and the improvement of MT-FwFM over FwFM is due to that MT-FwFM is able to learn different field interaction strengths for different conversion types, which will be described in details in Section~\ref{sec:analysis}.

%The AUC of ROC of different models on data set of each conversion type can be found in Table~\ref{table:mt-fwfm-vs-others-per-conversion-type} in appendix. MT-FwFM get the best performance on training data set on all conversion types. On the test data set, MT-FwFM get the best performance on 3 types: Lead, View Content and Sign Up, while FwFM performs better on Purchase.

\section{Study of Learned Field Interaction Effects for Different Conversion Types}
\begin{figure*}[ht]
  \captionsetup{width=.8\linewidth}
  \begin{subfigure}[t]{0.8\columnwidth}
    \centering
    \captionsetup{width=.8\linewidth}
    \includegraphics[width=\columnwidth]{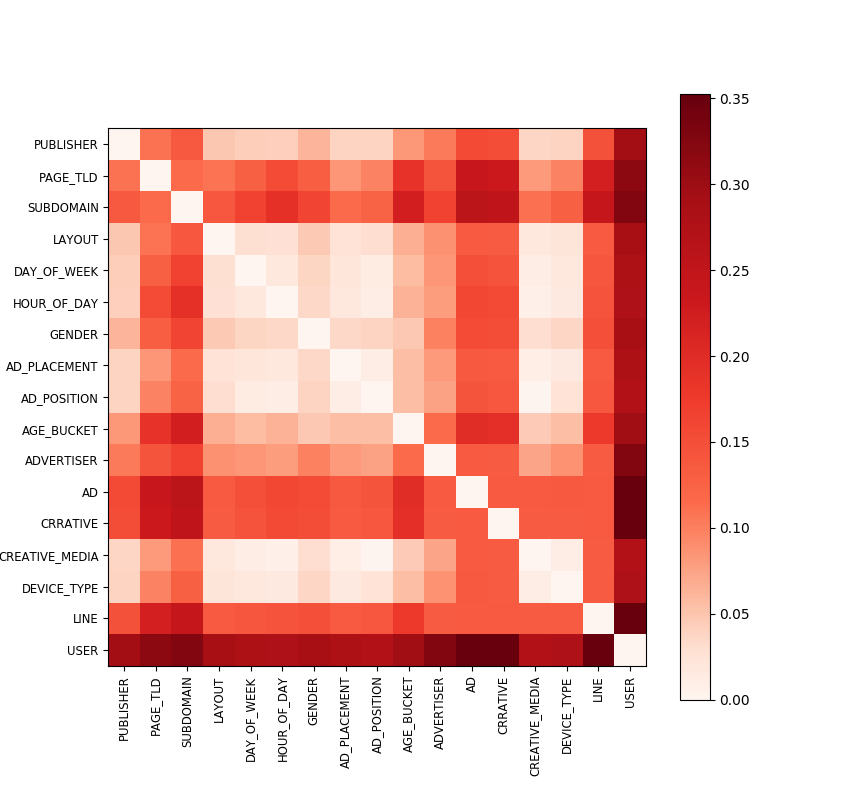} 
    \caption{Lead}
    \label{subfig:heat-map-mi-lead}
    \vspace{4ex}
  \end{subfigure}%%
  \begin{subfigure}[t]{0.8\columnwidth}
    \centering
    \captionsetup{width=.8\linewidth}
    \includegraphics[width=\columnwidth]{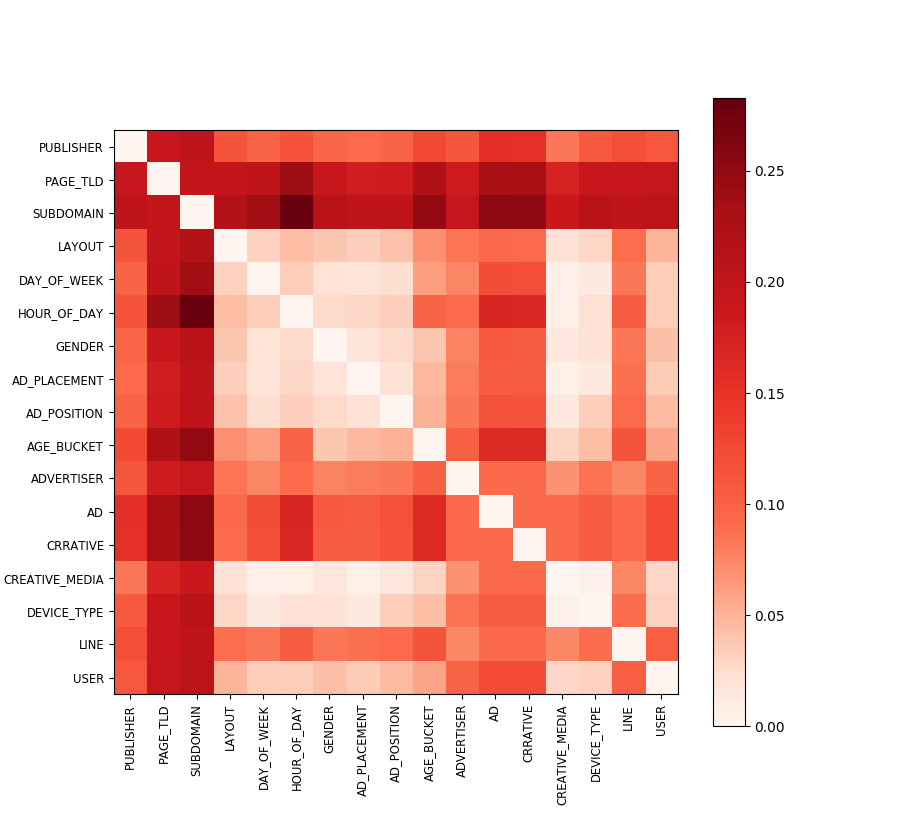} 
    \caption{View Content}
    \label{subfig:heat-map-mi-view-content}
    \vspace{4ex}
  \end{subfigure} 
  \begin{subfigure}[t]{0.8\columnwidth}
    \centering
    \captionsetup{width=.8\linewidth}
    \includegraphics[width=\columnwidth]{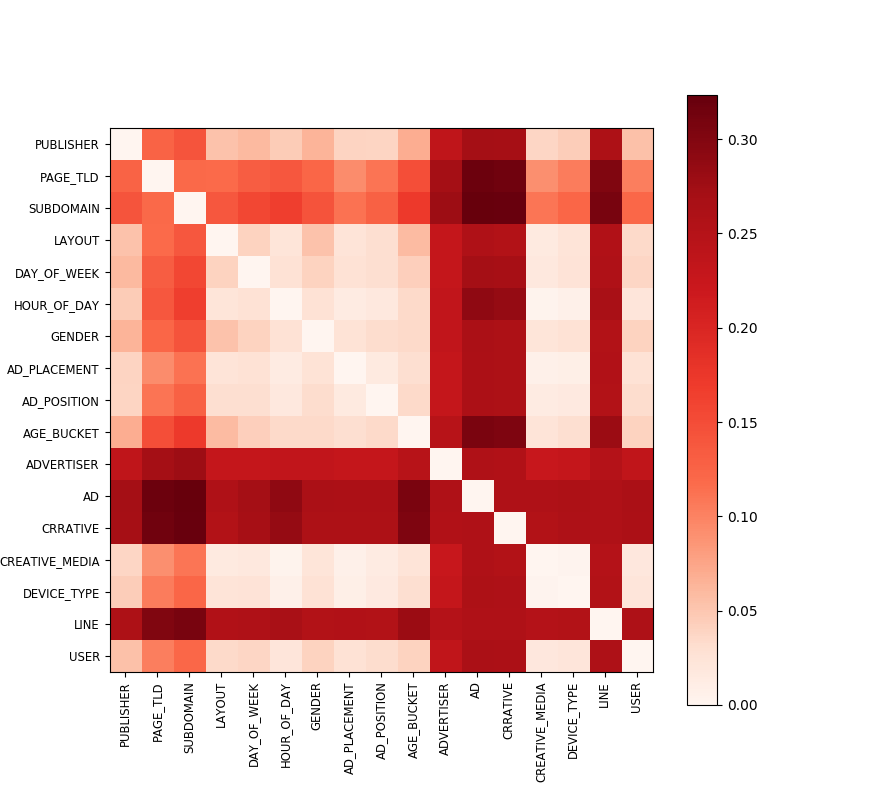} 
    \caption{Purchase}
    \label{subfig:heat-map-mi-purchase}
    \vspace{4ex}
  \end{subfigure}%% 
  \begin{subfigure}[t]{0.8\columnwidth}
    \centering
    \captionsetup{width=.8\linewidth}
    \includegraphics[width=\columnwidth]{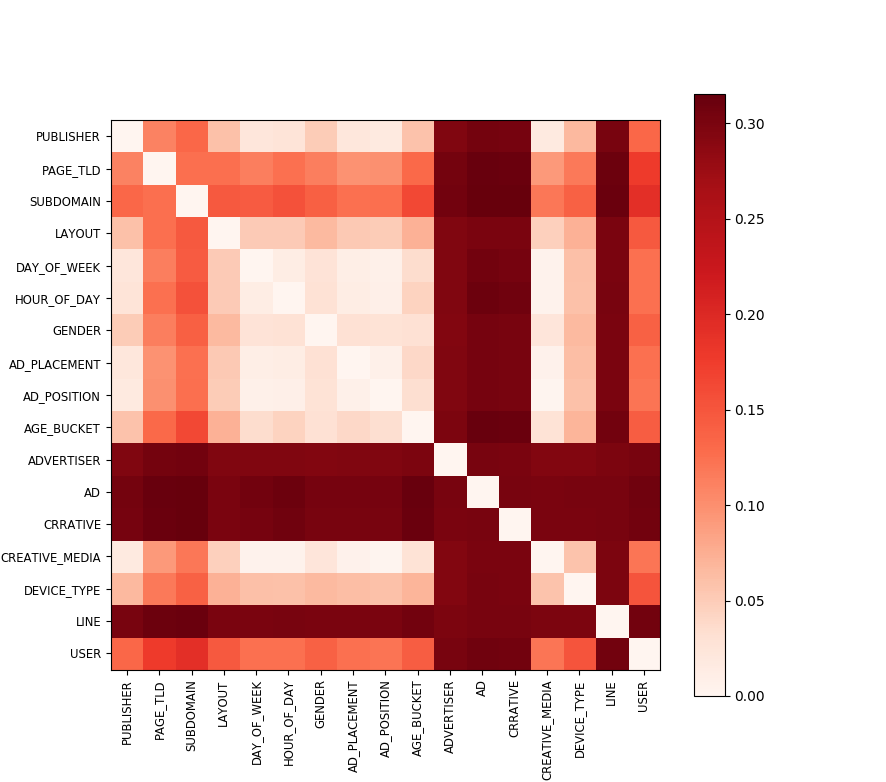} 
    \caption{Sign Up}
    \label{subfig:heat-map-mi-sign-up}
    \vspace{4ex}
  \end{subfigure} 
  \caption{Heat maps of mutual information between field pairs and each type of conversion.}
  \label{fig:heat-map-mi-different-conv-type}
\end{figure*}

In this section, we analyze MT-FwFM in terms of its ability to capture different field interaction effects for different conversion types. As described in Section~\ref{sec:main_interaction_effect}, the field interaction effects are measured by the the mutual information between a field pair $(F_p, F_q)$ and the conversion of each type, \textit{i.e.}, $MI^t((F_p,F_q),Y)$. Figure~\ref{fig:heat-map-mi-different-conv-type} presents the visualization of these field interaction effects by heat maps.

%We visualize it by a heat map, as shown in Figure~\ref{fig:heat-map-mi-different-conv-type}.

%the mutual information between field pairs and conversions for different conversion types. In Section~\ref{sec:learned} we show that the learned task-specific field interaction weights $r_{F_k, F_l}^t$ are highly correlated with mutual information.

\begin{figure*}[ht]
  \captionsetup{width=.8\linewidth}
  \centering
  \begin{subfigure}[t]{0.8\columnwidth}
    \centering
    \captionsetup{width=.8\linewidth}
    \includegraphics[width=\columnwidth]{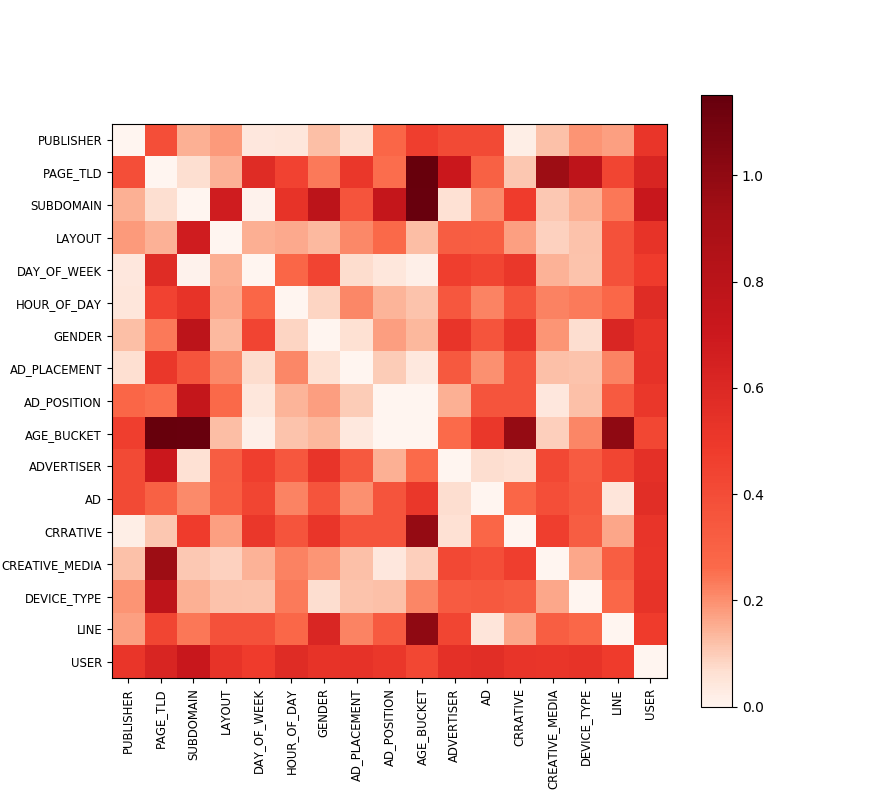} 
    \caption{Lead}
    \label{subfig:heat-map-mt-fwfm-r-lead}
    \vspace{4ex}
  \end{subfigure}%%
  \begin{subfigure}[t]{0.8\columnwidth}
    \centering
    \captionsetup{width=.8\linewidth}
    \includegraphics[width=\columnwidth]{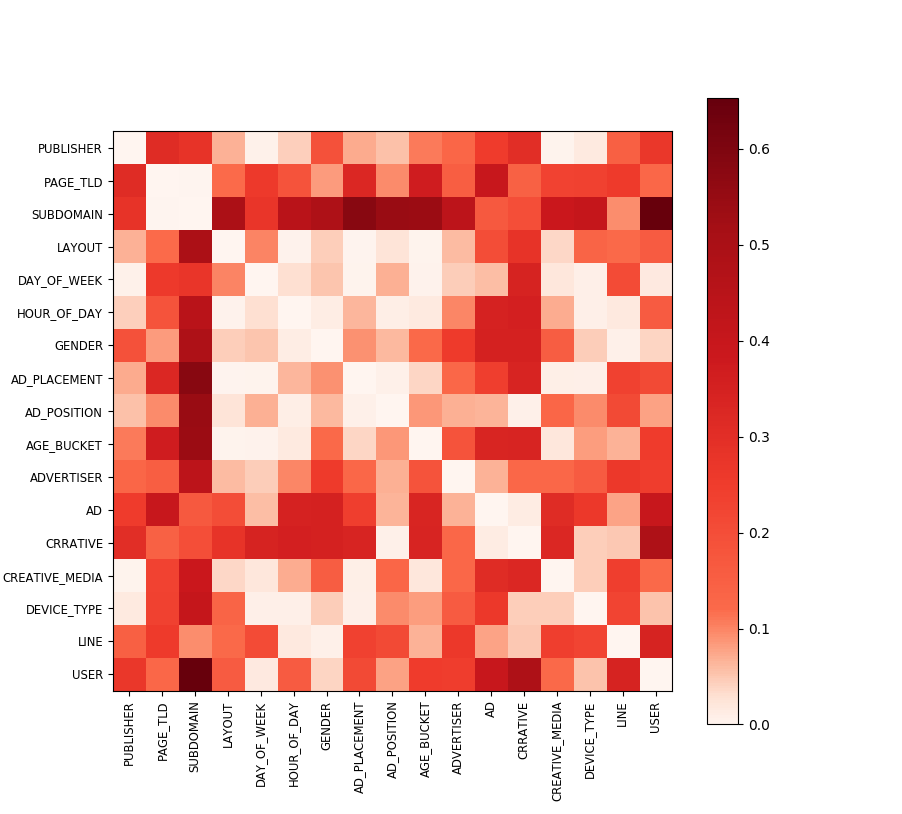} 
    \caption{View Content}
    \label{subfig:heat-map-mt-fwfm-r-view-content}
    \vspace{4ex}
  \end{subfigure} 
  \begin{subfigure}[t]{0.8\columnwidth}
    \centering
    \captionsetup{width=.8\linewidth}
    \includegraphics[width=\columnwidth]{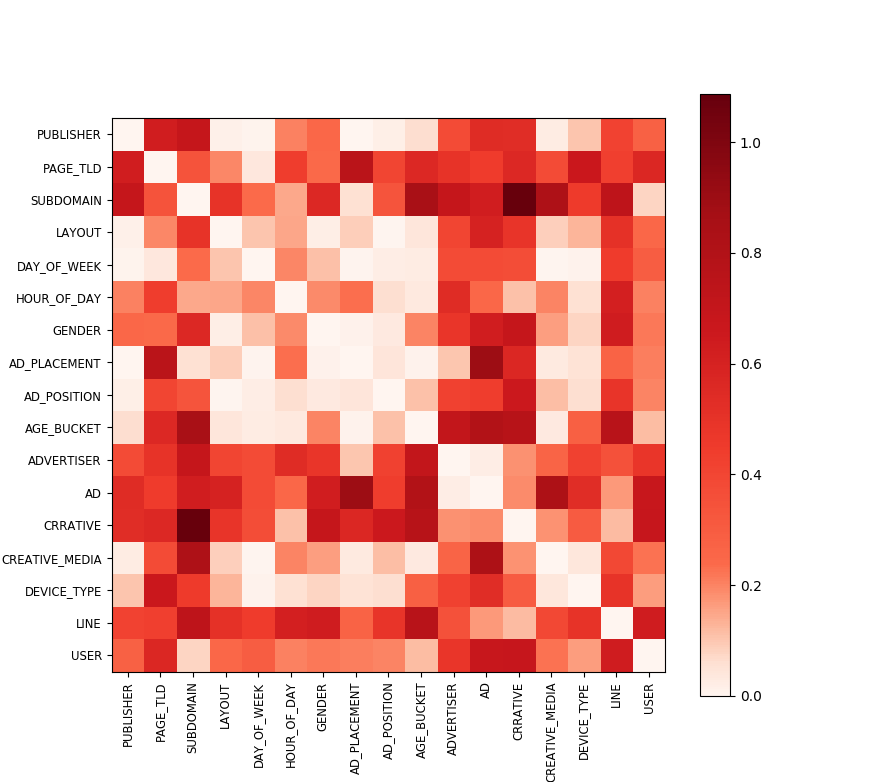} 
    \caption{Purchase}
    \label{subfig:heat-map-mt-fwfm-r-purchase}
    \vspace{4ex}
  \end{subfigure}%% 
  \begin{subfigure}[t]{0.8\columnwidth}
    \centering
    \captionsetup{width=.8\linewidth}
    \includegraphics[width=\columnwidth]{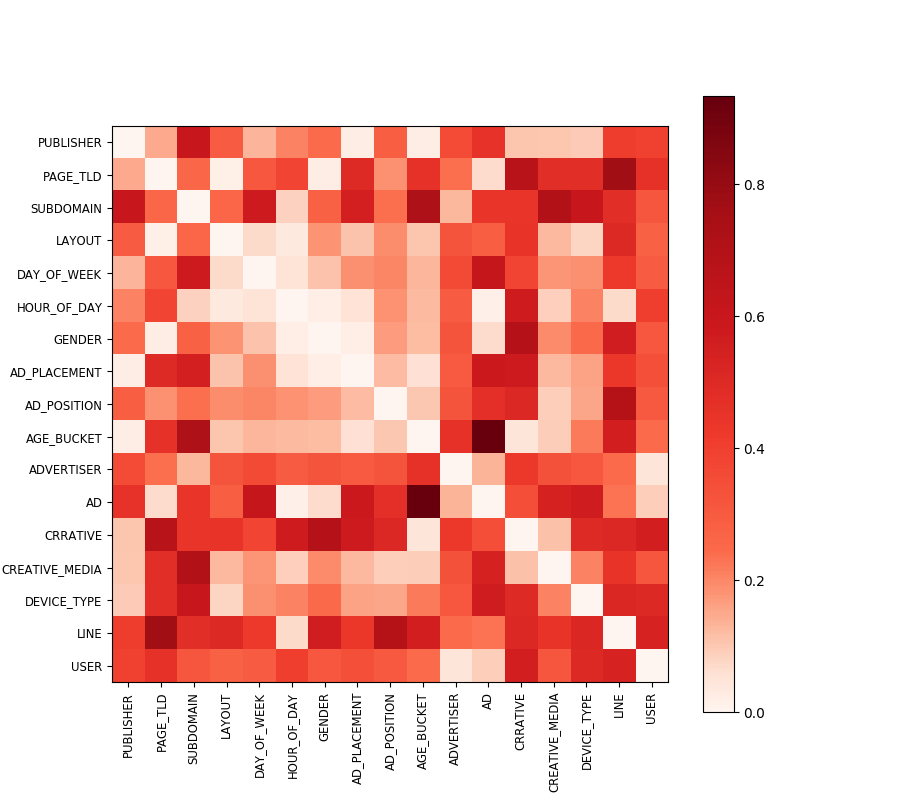} 
    \caption{Sign Up}
    \label{subfig:heat-map-mt-fwfm-r-sign-up}
    \vspace{4ex}
  \end{subfigure} 
  \caption{Heat maps of learned field interaction effects from MT-FwFM, \textit{i.e.}, $|\bm{r}^t_{F_k, F_l}|$ for different conversion types.}
  \label{fig:heat-map-mt-fwfm-r}
\end{figure*}

The difference among the four heat maps in Figure~\ref{fig:heat-map-mi-different-conv-type} illustrates how field interaction effects vary among different conversion types. 
For \textit{Lead}, \textit{User\_ID} has very strong interaction effects with almost all other fields, especially with \textit{Page\_TLD}, \textit{Subdomain}, \textit{Ad} and \textit{Creative}. 
For \textit{View Content}, field pairs containing publisher-side fields such as \textit{Page\_TLD} and \textit{Subdomain} have large mutual information in general.
For \textit{Purchase} and \textit{Sign Up}, we observe field pairs with advertiser-side fields, such as \textit{Advertiser}, \textit{Ad}, \textit{Creative} and \textit{Line}, have strong interaction effects with other fields.

To verify whether MT-FwFM captures the different patterns of field interaction effects among conversion types, we compare $MI^t((F_p,F_q),Y)$ with the learned field interaction effect between $F_p$ and $F_q$ on conversion type $t$, namely $|r^t_{F_p, F_q}|$. Here we only consider the magnitude of $r^t_{F_k,F_l}$, since either a large positive or negative value indicate a strong interaction effect. Figure~\ref{fig:heat-map-mt-fwfm-r} shows the heat maps of $|r^t_{F_p, F_q}|$ for all conversion types.

%If a model can capture the field interaction strength for each conversion type, we expect its learned field interaction effects for each type $t$, i.e., $r^t_{F_k, F_l}$ in MT-FwFM,  to be close to the mutual information of each type. To facilitate the comparison, in Figure~\ref{fig:heat-map-mt-fwfm-r} we plot the heat map of  $|r^t_{F_k,F_l|}$ for different conversion types. We take the absolute values of $r^t_{F_k,F_l}$ since both large positive or negative values denote a high interaction strength.

%We find that $|r^t_{F_k,F_l}|$ indeed learns different field interaction effects for different conversion types, and the learned field interaction effects match the mutual information for each type quite well. For \textit{Lead}, MT-FwFM learns that \textit{User}, \textit{Page\_TLD}, \textit{Subdomain} have in general stronger interaction effects with other fields. For \textit{View Content}, the publisher side fields, \textit{e.g.}, \textit{Publisher}, \textit{Page\_TLD}, \textit{Subdomain} are learned to have high interaction effects with other fields. It also learns that some field pairs that consist of both publisher side fields and advertiser fields, such as (\textit{Subdomain}, \textit{Ad}), (\textit{Page\_TLD}, \textit{Ad}) have very strong interaction effects. For \textit{Purchase} and \textit{Sign Up}, it learns that advertiser fields, \textit{e.g.}, \textit{Ad}, \textit{Creative} have in general strong interaction effects with other fields.

According to the comparison between Figure~\ref{fig:heat-map-mi-different-conv-type} and Figure~\ref{fig:heat-map-mt-fwfm-r}, the learned field interaction effects $|r^t_{F_k,F_l}|$ have similar pattern with their mutual information for each conversion type. 
In general, Figure~\ref{fig:heat-map-mt-fwfm-r} looks like a pixelated version of Figure~\ref{fig:heat-map-mi-different-conv-type}. 
For \textit{Lead}, MT-FwFM successfully captures that \textit{User\_ID} have strong interaction effects with other fields. 
For \textit{View Content}, field pairs including the publisher-side fields, \textit{e.g.}, \textit{Publisher}, \textit{Page\_TLD}, and \textit{Subdomain} generally have large magnitude of $|r^t_{F_k,F_l}|$. 
For \textit{Purchase} and \textit{Sign Up}, advertiser-side fields, \textit{e.g.}, \textit{Advertiser}, \textit{Ad}, \textit{Creative} and \textit{Line} have in general large $|r^t_{F_k,F_l}|$ with other fields.

\label{sec:analysis}

\section{Related Work}
%There has been lots of work in the literature on modeling clicks in display advertising. 
%\textcolor{red}{This would usually go somewhere before the results, usually after the introduction. No idea if TechPulse is more liberal on this.}

There has been lots of work in the literature on click and conversion prediction in online advertising. Research on click prediction focus on developing various models, including Logistic Regression (LR) ~\cite{richardson2007predicting,chapelle2015simple,mcmahan2013ad}, Polynomial-2 (Poly2) ~\cite{chang2010training}, tree-based models~\cite{he2014practical}, tensor-based models~\cite{rendle2010pairwise}, Bayesian models~\cite{graepel2010web}, Field-aware Factorization Machines (FFM) ~\cite{juan2016field,juan2017field}, and Field-weighted Factorization Machines (FwFM)~\cite{pan2018field}. Recently, deep learning for CTR prediction also attracted a lot of research attention~\cite{cheng2016wide,zhang2016deep,qu2016product,guo2017deepfm,shan2016deep,he2017neural,wang2017deep}.

For conversion prediction, \cite{lee2012estimating} present an approach to estimate conversion rate based on past performance observations along data hierarchies. \cite{chapelle2015simple} and \cite{agarwal2010estimating} propose a logistic regression model and log-linear model for conversion prediction, respectively. \cite{rosales2012post} provides comprehensive analysis and proposes a new model for post-click conversion prediction. \cite{bagherjeiran2010ranking} proposes a ranking model that optimize the conversion funnel even for CPC (Cost-per-Click) campaigns. \cite{ji2017time} proposes a time-aware conversion prediction model. \cite{lu2017practical} describes a practical framework for conversion prediction to tackle several challenges, including extremely sparse conversions, delayed feedback and attribution gaps. Recently, there are also several work on modeling the delay of conversions~\cite{chapelle2014modeling,yoshikawa2018nonparametric}. 

Multi-Task Learning (MTL)~\cite{caruana1998multitask} has been used successfully across multiple applications, from natural language processing~\cite{collobert2008unified}, speech recognition~\cite{deng2013new}, to computer vision~\cite{girshick2015fast}. MTL is also applied to online advertising in  ~\cite{ahmed2014scalable} to model clicks, conversions and unattributed conversions. In ~\cite{ma2018entire} the authors proposes a multi-task model to solve the tasks of click prediction and click-through conversion prediction jointly.
\label{sec:related_work}

\section{Conclusion}
%\st{In this paper, we first verify that the field interaction strengths are quite different for different conversion types. We then formulate the conversion prediction as a Multi-Task Learning problem and propose Multi-Task Field-weighted Factorization Machines(MT-FwFM) to resolve these tasks simultaneously.} \textcolor{blue}{In this paper we have introduced Multi-Task Field-weighted Factorization Machines(MT-FwFM), the first multi-task learning method specialized for conversion prediction. } Our experimental results show that MT-FwFM can outperform several state-of-the-art models, including logistic regression(LR), factorization machines(FM) and field-weighted factorization machines(FwFM). \st{Finally, comprehensive analysis verifies that MT-FwFM indeed learns different field interaction strengths for different conversion types.} \textcolor{blue}{Also, we have shown how MT-FwFM is able to to learn different field interactions for different conversion types by calculating the mutual information for some key parameters.} There are many potential directions for future research. \st{For example, it's interesting to involve more tasks to the current model, including click prediction and unattributed conversion prediction. Another direction is to build a unified deep neural networks to resolve the multi-task learning problem for conversion prediction.} \textcolor{blue}{To name a few, we could add click and unattributed conversion prediction as tasks, or make our current MT-FwFM model deep by adding pooling layers and others.}

In this paper, we formulate conversion prediction as a Multi-Task learning problem and propose Multi-Task Field-weighted Factorization Machines (MT-FwFM) to learn prediction models for multiple conversion types jointly. The feature representations are shared by all tasks while each model has its specific parameters, providing the benefit of sharing information among different conversion prediction tasks. Our extensive experiment results show that MT-FwFM outperforms several state-of-the-art models, including Factorization Machines (FM) and Field-weighted Factorization Machines (FwFM). We also show that MT-FwFM indeed learns different field interaction effects for different conversion types. There are many potential directions for future research. To name a few, we could involve more tasks to the current model, including predicting clicks or non-attributed conversions, or build a deep neural network (DNN) on top of MT-FwFM to better solve these tasks.
\label{sec:conclusion}

\bibliographystyle{ACM-Reference-Format}
\bibliography{sample-base}


\end{document}